\documentclass[letterpaper, 10 pt, conference]{ieeeconf} 
\IEEEoverridecommandlockouts

\usepackage{cite, balance}
\usepackage{amsmath,amssymb,amsfonts}
\usepackage{algorithmic}
\usepackage{graphicx}
\usepackage{textcomp}
\usepackage{xcolor}
\DeclareMathAlphabet{\mathpzc}{OT1}{pzc}{m}{it}

\DeclareMathAlphabet\mathbfcal{OMS}{cmsy}{b}{n}

\newcommand{\asy}{\mathit{as}} %asymetric part

\newcommand{\Ad}{\mathit{Ad}} %Adjoint
\newcommand{\Rot}{\mathbf{R}} %Rotation matrix
\newcommand{\pos}{\mathbf{p}}
\newcommand{\homTr}{\mathbf{H}} %Tranformation matrix
\newcommand{\Costif}{\mathbf{G}} %Costiffness
\newcommand{\Stif}{\mathbf{K}} %stiffness
\newcommand{\Damp}{\mathbf{B}}%damping
\newcommand{\Jacobi}{\mathbf{J}}%Jacobi
\newcommand{\Id}{ \mathbf{I}}%Identity  
\newcommand{\jp}{\mathbf{q}}%joint positon vector
\newcommand{\jv}{\mathbf{\dot{q}}}%joint velocity vector
\newcommand{\tq}{\boldsymbol{\tau}}%joint torque vector
\newcommand{\wrench}{\mathbf{w}}%wrench

\newcommand{\potEN}{U}%inertia Tensor  
\newcommand{\kinEN}{T}
\newcommand{\powmot}{P_{\text{motion}}}
\newcommand{\entot}{E_{\text{total}}}
\newcommand{\powmotUP}{\overline{P}_{\text{motion}}}
\newcommand{\entotUP}{\overline{E}_{\text{total}}}

\newcommand{\powtask}{P_{\text{task}}}
\newcommand{\powtankLOW}{\underline{P}_{\text{tank}}}
\newcommand{\entank}{E_{\text{tank}}}
\newcommand{\entankUP}{\overline{E}_{\text{tank}}}
\newcommand{\entankLOW}{\underline{E}_{\text{tank}}}
\newcommand{\SE}{\mathit{SE}} 

\newcommand{\REAL}{\mathbb{R}}%Realnumber

\newcommand{\Zax}{z\text{-axis}}

\usepackage{import}
\usepackage{tikz}
\usepackage{standalone}
\usepackage{pgfplots}	
\usepackage{siunitx}
\usepackage{adjustbox}
\let\labelindent\relax
\usepackage[inline]{enumitem}
\usepackage{hyperref}
\pgfplotsset{compat=1.16}
  \usetikzlibrary{plotmarks}
  \usetikzlibrary{arrows.meta}
  \usepgfplotslibrary{patchplots}
  \usepackage{grffile}
  \usepackage{amsmath}
\usepackage{cite}

\begin{document}
\title{\LARGE \bf Design of an Energy-Aware Cartesian Impedance Controller \\ for Collaborative Disassembly}

\author{Sebastian Hjorth$^{1}$, Edoardo Lamon$^{2}$, Dimitrios Chrysostomou$^{1}$, and Arash Ajoudani$^{2}$% 
\thanks{\noindent$^1$Dept. of Materials and Production, Aalborg University, Aalborg, Denmark. \tt\small sshj@mp.aau.dk}
\thanks{\noindent$^2$Human-Robot Interfaces and Interaction, Istituto Italiano di Tecnologia, Genoa, Italy. \tt\small edoardo.lamon@iit.it}
\thanks{\noindent This research was partly supported by EU's SMART EUREKA programme S0218-chARmER, Innovation Fund Denmark (Grant no. 9118-00001B), by the EU’s H2020-WIDESPREAD project no. 857061 "Networking for Research and Development of Human Interactive and Sensitive Robotics Taking Advantage of Additive Manufacturing -- R2P2", and by EU’s Horizon 2020 research and innovation programme under Grant Agreement No. 871237 (SOPHIA).}}
\maketitle
\thispagestyle{empty}
\pagestyle{empty}

\begin{abstract}
Human-robot collaborative disassembly is an emerging trend in the sustainable recycling process of electronic and mechanical products. It requires the use of advanced technologies to assist workers in repetitive physical tasks and deal with creaky and potentially damaged components. Nevertheless, when disassembling worn-out or damaged components, unexpected robot behaviors may emerge, so harmless and symbiotic physical interaction with humans and the environment becomes paramount. This work addresses this challenge at the control level by ensuring safe and passive behaviors in unplanned interactions and contact losses. 
The proposed algorithm capitalizes on an energy-aware Cartesian impedance controller, which features energy scaling and damping injection, and an augmented energy tank, which limits the power flow from the controller to the robot. The controller is evaluated in a real-world flawed unscrewing task with a Franka Emika Panda and is compared to a standard impedance controller and a hybrid force-impedance controller. The results demonstrate the high potential of the algorithm in human-robot collaborative disassembly tasks. 
\end{abstract}

\section{Introduction}~\label{sec:Intro}
\begin{figure}[t]
    \centering
    \adjustbox{width=0.75\columnwidth}{\import{Figures/Concept/}{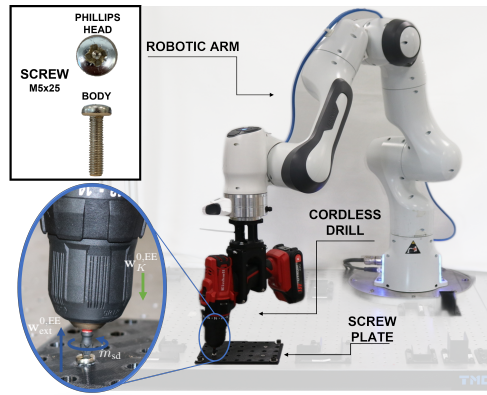}}
    \caption{Experimental setup: a manipulator, equipped with a cordless drill unfastens screws on the screw plate placed on the same workbench. The robot controller can drive the drill through a microcontroller.}
    \label{fig:unscrewing_experimental_setup}
    \vspace{-5mm}
\end{figure}

\noindent In the last decade, the European Union has started to promote the implementation of circular economy business models (CEBMs) across different manufacturing areas~\cite{eu2015ce,eu2020ea}.
CEBMs envision the adoption of take-back programs, efficient disassembly, and requalification processes~\cite{graedel2011ce}.
To make such business models financially, environmentally, and socially viable, companies must recover as many undamaged components as possible. However, various challenges, for example, high variability in the condition of post-use parts, poor information about returned products, high product complexity, increasing quality requirements on recovered materials and components, and pressure on costs and efficiency, strongly limit the wide exploitation of effective disassembly processes~\cite{tolio2017ce}. \\
\noindent To face the variability and uncertainties of the products' state~\cite{elo2014lcd}, human-in-the-loop solutions such as human-robot collaborative disassembly (HRCD) cells, in which humans and robots support each other to complete a given non-destructive disassembly task, started to be conceived~\cite{hjorth2022human} (an example for an HRCD cell can be seen in~\autoref{fig:unscrewing_experimental_setup}.).
Nonetheless, when it comes to physical human-robot interaction (pHRI), implementing such HRCD solutions demands high safety standards.
Due to their design, collaborative-enabled robots can attain control strategies that regulate the level of compliance ~\cite{zhao2022hybrid}, enabling them to interact safely with unstructured environments~\cite{pereira2022improving}. These features are of great importance in HRCD, which presents a high risk of unpredictable events occurring due to the variability and uncertainties of the product's state. In particular, when separating two sub-assemblies, the breakage of the fasting component could lead to a contact loss between the robot's end-effector and the component, which, in turn, might result in damaged components, tools and, in the worst case, human harm.  

\noindent A common fastening method that does not require destructive disassembly is screwing. However, since the product has reached its end of life, the screw condition can significantly affect the disassembly process~\cite{vongbunyong2015daas}. 
Several control strategies for HRCD have been proposed in the literature, but only a few could deal with the realistic flawed conditions mentioned above. 
The work done by~\cite{chen2014unscrewing} focuses on the unscrewing of components on lithium-ion car batteries and the mechanism for changing the tool bit autonomously. It concluded that complained control schemes (i.e. impedance controller) are necessary to enable safe and direct pHRI.
The work presented in~\cite{li2020unscrewing,huang2021discell} focuses on the unfastening of screws with an external hexagonal-shaped head utilizing a KUKA LBR iiwa. Unscrewing is achieved with the help of a standard Cartesian impedance control scheme in combination with a custom nut-runner, which encloses the hexagonal-shaped screw head.
Another unscrewing strategy for screws with a hexagonal shaped screw head was proposed in~\cite{rastegarpanah2021unscrewing}, where screw location and orientation are detected and the unscrewing makes use of a standard Cartesian impedance controller and a conventional 2-fingers gripper.
In~\cite{mironov2018haptics}, an unscrewing robotic system for the automatic disassembly of electronic devices was developed. The work investigated force and torque profiles used by humans when unscrewing Phillips and internal hex screws, to design a control strategy and tool that minimizes slippage. The proposed strategy was deployed on a position-controlled UR3 equipped with a force/torque sensor in combination with a purpose-built screwdriver with passive compliance along the tool $z$-axis. 
The drawback of using a position-controlled robot is that in case of collision, the robot's only safety mechanism is the emergency stop, which can result in a dangerous quasi-static contact scenario (e.g., clamping).
\noindent Nevertheless, none of the above presented approaches considers the robot's behavior when an unpredicted faulty situation occurs during the unscrewing due to either a broken screw or failed engagement during the unscrewing process.
In this context, observing, monitoring, and limiting the amount of energy and power flow that the controller is allowed to inject into the manipulator is crucial to a successful interaction. Additionally, the energy exchange between the manipulator and its environment would result in a safer task execution~\cite{lachner2021energybudget}, not only by ensuring a passive behavior, but also by reducing the controller's action in potential faults.
Different control schemes that could allow the limitation of the energy and power flow are presented in~\cite{stefano2001ipc, raiola2018energy, lachner2021energybudget}.
Therefore, the presented formalism will focus on an energy-aware control strategy with the aim to tackle the aforementioned challenges. The formalism is an extension of an energy-aware Cartesian impedance controller presented in~\cite{raiola2018energy,hjorth2020energy} in combination with a task-based energy tank and a power flow regulation mechanism proposed by~\cite{shahriari2019power}. Moreover, the controller's performance is evaluated and compared with the controller proposed in the literature through a set of experiments. These experiments assess the controller's ability to handle external disturbances and minimize the impact force and energy exchange with the environment.
To summarize, the novel contributions of the manuscript are the following:
\begin{itemize}
\item The design of an energy-aware Cartesian impedance controller that uses a global energy tank with power limitation.
\item An experimental comparison of the proposed energy-aware Cartesian impedance controller with a hybrid force-impedance controller~\cite{shahriari2019power}, and a standard Cartesian impedance controller~\cite{ott2008cartesian}, in a faulty unscrewing task with a stripped screw head. 
\end{itemize}
\vspace{-2mm}
\section{Problem Statement}\label{sec:problem}
\begin{figure}[t]
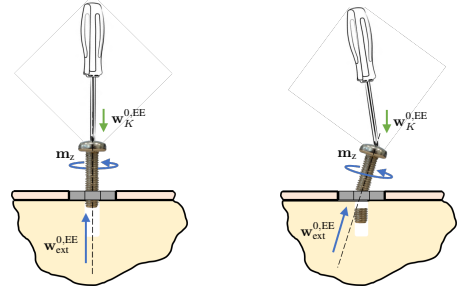

    \centering
      \adjustbox{width=0.27\columnwidth}{\import{Figures/Concept/}{unscrewing_g.tex}}
    \hspace{1cm}
     \adjustbox{width=0.27\columnwidth}{\import{Figures/Concept/}{unscrewing_b.tex}}
    \caption{Illustration of two different unscrewing scenarios. Left-hand side: a successful unscrewing operation. Right-hand side: a possible scenario for contact loss due to the screw shaft breakage during the unscrewing operation.}
    \label{fig:screwing_problem}
    \vspace{-6mm}
\end{figure}

\noindent In this paper, we examine an HRCD task with a redundant manipulator, focusing on the behavior in case of contact loss during the unscrewing. Such a scenario can occur due to a stripped screw head, corrosion, or screw damaged inside that is not detectable by external inspection (e.g., damaged threads or broken shaft) or due to a pHRI.
The generalized approach for the unscrewing operation is to generate enough momentum $\mathbf{m_\text{z}}$ on the screw so that the screw rotates around its longitudinal axis, thus generating a vertical upward force $\wrench^{0,\text{EE}}_{\text{ext}}$.
In order to ensure the engagement of the screwdriver bit during the unscrewing process, it is crucial to apply a force ${\wrench^{\text{0},\text{EE}}_K}$ onto the screw throughout the entire unscrewing process. Common approaches to generate such force are standard Cartesian impedance and hybrid Cartesian force-impedance controllers.
In case of a contact loss during the unscrewing process due to one of the scenarios mentioned above, the force applied at the screwdriver might cause the bit to hit the screwed product, damage the two components of the screwed product or harm the operator.
A graphical visualization of a successful and a failed unscrewing operation is depicted in \autoref{fig:screwing_problem}.
\vspace{-2mm}
\section{Methodology}~\label{sec:methods}
\vspace{-6mm}
\subsection{Energy Aware-Impedance Controller}\label{sub:control}
\noindent This control scheme presents a method that autonomously counteracts the possible non-linear behavior of the impedance model in contact scenarios~\cite{ott2008cartesian}, with the help of the energy shaping and damping injection techniques, firstly introduced in~\cite{stefano1998pdh}. These concepts are utilized to monitor and limit the total energy of the system, as well as the power of the robot.
In~\cite{raiola2018energy,hjorth2020energy}, the same control strategy was successfully implemented on a kinematically redundant manipulator, whose control torques $\tq^\top_{\text{Control}}\in\REAL^n$ are defined in a quasi-static condition as \vspace{-2mm}
\begin{equation}\label{eq:tau_imp} 
    \tq^\top_{\text{Control}} = \tq^\top_{\text{Spring}} - \tq^\top_{\text{Damp}},
\end{equation} \vspace{-0.5mm}
where $\tq^\top_{\text{Spring}}\in\REAL^n$ and $\tq^\top_{\text{Damp}}\in\REAL^n$ are respectively generated by Cartesian springs and dampers. For a more detailed explanation of the notation hereafter, please refer to~\cite{hjorth2020energy}. The torques $\tq^\top_{\text{Spring}}$ are generated by the elastic wrench ${\wrench^{\text{EE},\text{EE}}_K}^\top\in se^{\ast}(3)$, which can be expressed as \vspace{-0.5mm}%
\begin{equation} \label{eq:WrenchEE_K}
    {\wrench^{\text{EE},\text{EE}}_K}^\top = \begin{bmatrix}
    {\mathbf{f}^{\text{EE},\text{EE}}_K}^\top  \\ 
    {\mathbf{m} ^{\text{EE},\text{EE}}_K}^\top  \\
    \end{bmatrix} = \begin{bmatrix}
    	\Stif_t & \Stif_c \\ 
    	\Stif^\top_c & \Stif_r \\
    	\end{bmatrix} \Delta{\boldsymbol{\eta}}
\end{equation}\vspace{-0.5mm}
where $\Delta{\boldsymbol{\eta}}\in se(3)$ describes the infinitesimal body twist displacement~\cite{stefano2001ipc}, the diagonal matrices $\Stif_{r}\in\REAL^{3\times3}$, $\Stif_{t}\in\REAL^{3\times3}$ hold the stiffness values for the rotational, translation springs, and $\Stif_{c}\in\REAL^{3\times3}$ describes the decoupling between these two terms. Force ${\mathbf{f}^{\text{EE},\text{EE}}_K}\in\REAL^{1\times3}$ and momentum ${\mathbf{m} ^{\text{EE},\text{EE}}_K}\in\REAL^{1\times3}$ can be formulated in terms of energy by basing their formulation on the end-effector's current transformation $\homTr^{0}_{\text{EE}}\in\SE(3)$ and its desired goal transformation $\homTr^{0}_{d}\in\SE(3)$ in the following way:
\vspace{-0.15mm}
\begin{equation}\label{eq:fm_ee_k}\begin{aligned}\widehat{\mathbf{f}}^{\text{EE},\text{EE}}_K =& -\Rot^{\text{EE}}_d \asy\big(\Costif_t \widehat{\mathbf{p}}^d_{\text{EE}}\big)\Rot^d_{\text{EE}}-\asy\big(\Costif_t \Rot^{\text{EE}}_d  \widehat{\mathbf{p}}^d_{\text{EE}} \Rot^d_{\text{EE}} \big) \\& -2\asy\big(\Costif_c \Rot^d_{\text{EE}}\big) \\ \widehat{\mathbf{m}}^{\text{EE},\text{EE}}_K =& -2\asy\big(\Costif_r  \Rot^d_{\text{EE}}\big)- \asy\big( \Costif_t  \Rot^{\text{EE}}_d  \widehat{\mathbf{p}}^d_{\text{EE}} \widehat{\mathbf{p}}^d_{\text{EE}} \Rot^d_{\text{EE}} \big) \\& -2\asy\big(\Costif_c \widehat{\mathbf{p}}^d_{\text{EE}} \Rot^d_{\text{EE}}\big),
    \end{aligned}
\end{equation}\vspace{-0.65mm}
with $\pos^d_{\text{EE}}\in\REAL^3$, $\Rot^{\text{EE}}_d\in SO(3)$ describe the translation and rotation between the end-effector's current and its desired configuration. $\Costif_{r,t,c}\in\REAL^{3\times3}$ are co-stiffnesses of the rotational spring, translational spring and coupling term \cite{thadel_phd,hjorth2020energy,raiola2018energy} and $\asy()$ represents the asymmetric part of the matrix.
\noindent Safety is dealt with two different controller features:
\begin{enumerate*}[label={(\roman*)}]
    \item monitoring the total amount of energy stored in the system with the help of energy scaling and 
    \item limiting the power of the system with the damping injection method, if necessary.
\end{enumerate*}
The energy scaling method enforces a limit on the total energy of the system based on an energy-based safety metric $\entotUP$.
The total energy stored in the system can be expressed as $\entot = \kinEN_{\text{total}} + \potEN_{\text{total}}$ where $T_{\text{total}}\in\REAL$ is the kinetic co-energy and $\potEN_{\text{total}}\in\REAL$ the potential energy due to spatial springs~\cite{thadel_phd}.
In the event of a pHRI that results in a displacement of the end-effector, such that the statement $\entot > \entotUP$ becomes true, one computes the following scaling parameter:
\vspace{-1.25mm}
\begin{equation}\label{eq:energy_lambda}
    \centering
    \lambda = 
    \begin{cases} 
    1    &   \text{if } \entot \leqslant \entotUP  \\
    \frac{\entotUP-\kinEN}{\potEN_{\text{total}}}& \text{otherwise.}
    \end{cases}
\end{equation}\vspace{-0.05mm}
As seen in~\cite{raiola2018energy} $\potEN_{\text{total}}$ is
proportional to the co-stiffness $\Costif_{r,t,c}$. Therefore, by scaling $\Costif_{r,t,c}$ with $\lambda\in\REAL$ in the following way $\Costif_{r,t,c}\boldsymbol{\leftarrow} \lambda\Costif_{r,t,c}$, the wrench generated by the springs ${\wrench^{0,\text{EE}}_K}^\top$ is directly affected. This results in the motion generating torques from the Cartesian springs $\tq^\top_{\text{Spring}}=\Jacobi(\jp){\wrench^{0,\text{\text{EE}}}_{\text{K}}}^\top$, where ${\wrench^{0,\text{EE}}_K}^\top=\Ad^\top_{\homTr^{\text{EE}}_0}{\wrench^{\text{EE},\text{EE}}_K}^\top$ with $\Ad^\top_{\homTr^{\text{EE}}_0}\in\REAL^{6\times 6}$ being the adjoint coordinate transformation.
However, as the energy of the robot is manipulated directly, it is vital to ensure the passivity of the system. The enforcement of the passivity will be discussed in Section~\ref{sub:Energy-tank}.
After limiting the total energy of the robot, the robot's power must also be overseen, as the power describes the instantaneous energy transferred when the robot makes contact with its environment. For this purpose, the damping injection method monitors the power resulting from the manipulator's motion $\powmot\in\REAL$:\vspace{-1.75mm}
\begin{equation}\label{eq:powermotion_modified} 
    \powmot =\big(\Jacobi(\jp)^\top {\wrench^{0,\text{EE}}_K}^\top - \Damp_{\text{init}}\jv\big)^\top\jv,
\end{equation}\vspace{-0.75mm}
with $\Damp_{\text{init}}\in\REAL^{n\times n}$ being the initial positive definite damping matrix. As soon as the robot starts moving towards a desired transformation $\homTr^0_d$, $\powmot$ is monitored. In the event of $\powmot$ exceeding the chosen power limit $\powmotUP\in\REAL$, the scaling parameter $\beta\in\REAL$ is calculated:
\begin{equation}\label{eq:beta_damp_modified}
    \beta = \begin{cases} 
             1    &   \text{if }\powmot \leqslant \powmotUP \\
            \frac{\big(\Jacobi^{0,0}_{\text{EE}}(\jp)^\top {\wrench^{0,\text{EE}}_K}^\top \big)^\top \jv - \powmotUP}{\jv^\top \Damp_{\text{init}}\jv}& \text{otherwise,}
        \end{cases}
\end{equation}
multiplying $\beta$ with the initial damping matrix $\Damp_{\text{init}}$, resulting in the damping term:\vspace{-0.25mm}
\begin{equation}\label{eq:tau_damp}
    \tq^\top_{\text{Damp}}= \beta\Damp_{\text{init}}\jv.
\end{equation}
\vspace{-0.25mm}
Hence, in the scenario in which $\powmot$ exceeds $\powmotUP$, the increase of $\beta$ has a direct effect on the damping term.
\subsection{Energy Tank Integration}\label{sub:Energy-tank}
\noindent The previously described methods ensure the robot's safety by manipulating the robot's energy; however, manipulating the system's energy can result in a passivity-violating behavior. Therefore, it is necessary to limit the amount of energy that the controller is allowed to inject into the system.
Furthermore, to this aim, latest research, such as~\cite{michel2022safety, shahriari2019power, ramuzat2022passive, gerlagh2021energysim, gerlagh2022energytask} also highlights that one has to limit the rate of energy injected by the controller.
The augmented energy tank described in~\cite{shahriari2019power} prevents the energy of the system from suddenly increasing within a single time step, thereby ensuring that the system stays stable. As mentioned in~\cite{lachner2022shaping}, the stored energy in an impedance-controlled robot can be expressed as a storage function $S = S_{\text{c}}+S_{\text{r}}\in\REAL$, which is composed of the storage functions of the controller and the robot.
$S_{\text{r}}$ is passive as the robot's energy is physically bounded from below~\cite{stefano2015earbook}.
This results in $S$ being passive if $S_{\text{c}}$ is passive with respect to the possible violating ports. 
The power flow of the possible violating ports with respect to the energy storage function $S_{\text{c}}$ can be expressed as:
\vspace{-0.75mm}
\begin{equation}
    \dot{S_{\text{c}}} + P_\text{dissipation} + P_\text{task} = 0,
\end{equation}
\vspace{-0.25mm}
where $P_\text{dissipation}\in\REAL$ is the power flow due to the dissipation, and $\powtask = -\wrench^{\text{0},\text{EE}}_K\dot{\mathbf{x}}\in\REAL$ describes the power demand of the task and $\dot{\mathbf{x}}\in\REAL^6$ denotes the spatial end-effector velocity.
The passivity of the overall system can be achieved by augmenting the storage function $S_{\text{c}}$ with an energy tank $\entank$ bounded by upper and lower bounds $\entankUP$/$\entankLOW$.
The power flow of the new storage function can be expressed as\vspace{-1.5mm}
\begin{equation}
    \dot{S_{\text{c}}} + \dot{E}_{\text{tank}} \leq 0, 
\end{equation}\vspace{-2.5mm}
where 
\begin{equation}
    \dot{E}_{\text{tank}} = \powtask.
\end{equation}
\vspace{-0.25mm}
However, as pointed out in~\cite{shahriari2019power}, a rapid increase of the system energy even with a bounding $\entank$ unstable behavior can occur. Therefore, the authors proposed to limit the positive power flow in the system through the following formalism:\vspace{-1mm}
\begin{equation}
    P_{\text{task}} = \begin{cases}
        \gamma\mathit{k}\powtask & \text{if } \powtask \leq 0 \\
        \mathit{j}\powtask & \text{otherwise},
    \end{cases}
\end{equation}
\vspace{-0.25mm}
where $\mathit{k}$ and $\mathit{j}$ ensure that the upper and lower bounds of $\entank$ are not violated. 
They are defined in such a way that the control system neither injects or takes out energy from the tank if the respective bound is reached.
\begin{equation}
\begin{aligned}
    \mathit{k} &= \begin{cases}
        0 & \text{if } \powtask \leq 0 \wedge \entank \leq \entankLOW \\
        1 & \text{otherwise}
    \end{cases}\\ 
      \mathit{j} &= \begin{cases}
        0 & \text{if } \powtask \geq 0 \wedge E_{\text{tank}} \geq \entankUP \\
        1 & \text{otherwise}
    \end{cases}
\end{aligned}
\end{equation}
Additionally, the rate at which the controller can inject energy into the system is limited by: 
\begin{equation}
  \gamma = \begin{cases}
    \frac{\powtankLOW} {\powtask} & \text{if } \powtask < \powtankLOW \leq 0\\
    1 & \text{otherwise}.
 \end{cases}
\end{equation}
Where $\gamma$ is defined as a ratio between the maximal allowed power flow $\powtankLOW$ and the originally calculated power flow $\powtask$ from the controller to the system.
Integrating the above-described energy tank dynamic for the previously presented energy-aware Cartesian impedance controller, the energy scaling variable can be reformulated as 
\begin{equation}
 \lambda = 
    \begin{cases} 
    1    &   \text{if } \entot \leqslant \entotUP \wedge \mathit{k} \neq 0  \\
    \lambda(t-1) & \text{if }    \mathit{k} = 0 \wedge \powtask \leq 0\\
    \frac{\entotUP -\kinEN}{\potEN_{\text{total}}}& \text{otherwise.}
    \end{cases}
\end{equation}
In case the energy tank is empty ($\mathit{k}=0$), $\lambda$ is hindered from increasing again after an interaction occurs, thereby keeping the $\Costif_{r,t,c}$ constant, which results in a standard Cartesian impedance controller with constant gains. This will reduce the controller's performance; however, it does not hinder $\lambda$ from being further scaled down in order to ensure the safety metric $\entotUP$ is not violated. Additionally, to ensure that $\lambda$ does not increase when the energy tank is drained, it is important to ensure that $\lambda$ does not rise too fast and result in an unstable behavior of the manipulator.  Therefore, as previously mentioned, it is necessary to limit the energy that can be drained from the tank ($\powtask$). 
Hence, applying the constraint $\gamma$, the power flow from the tank to the system results in the following ${\tq}^\top_{\text{Spring}}$:
\begin{equation}\label{eq:tau_motion_new}
   {\tq}^\top_{\text{Spring}}= \gamma{\Jacobi}^\top(\jp) {\wrench^{0,\text{EE}}_K}^\top
\end{equation}
\noindent After limiting the amount of energy that can be injected into the system through the energy scaling variable $\lambda$, the damping injection terms must be modified as follows:
\vspace{-2mm}
\begin{equation}
    \powmot=\Bigg(\gamma\bigg(\Jacobi(\jp)^\top {\wrench^{0,\text{EE}}_K}^\top\bigg) - \Damp_{\text{init}}\jv\Bigg)^\top\jv
\end{equation}
\begin{equation}\label{eq:beta_damp}
    \beta = \begin{cases} 
             1    &   \text{if }\powmot \leqslant \powmotUP   \\
            \frac{\Bigg(\gamma\big(\Jacobi(\jp)^\top {\wrench^{0,\text{EE}}_K}^\top \big)\Bigg)^\top \jv - \powmotUP}{\jv^\top \Damp_{\text{init}}\jv}& \text{otherwise.}
        \end{cases}
\end{equation}
As $\gamma$ restricts the power flow to the system, it also has a direct effect on $\powmot$ resulting in the following control law:
\vspace{-0.5mm}
\begin{equation}
    \tau_{\text{Control}}^\top = \gamma{\Jacobi}^\top(\jp) {\wrench^{0,\text{EE}}_K}^\top - \beta\Damp_{\text{init}}\jv.
\end{equation}
\section{Experimental Results}\label{sec:Experiment}
\begin{figure*}[t]
\centering
\adjustbox{width=\textwidth}{\import{Figures/Controller/}{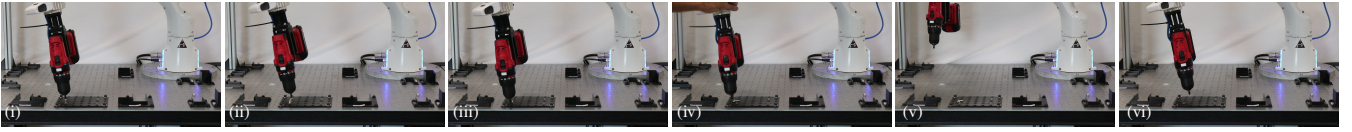}}
	\caption{Snapshots of the experiment with the energy-aware impedance controller. A video of the experiments is available in the multimedia extension and at  \href{https://www.youtube-nocookie.com/embed/SgYFHMlEl0k}{https://www.youtube-nocookie.com/embed/SgYFHMlEl0k}. From left to right: (i) unscrewing, (ii) contact loss, (iii) impact with table, (iv) pHRI, (v) end of the pHRI, (vi) second impact with table. }
	\label{fig:snapshots}
 \vspace{-3mm}
\end{figure*}

\noindent The energy-aware Cartesian impedance controller was tested in a proof-of-concept unscrewing experiment with a stripped screw by comparing its performance with state-of-art approaches, i.e., a standard Cartesian impedance controller~\cite{ott2008cartesian} and a Cartesian hybrid force-impedance controller with power limitation~\cite{shahriari2019power}. The setup consists of a Franka Emika Panada manipulator equipped with a cordless drill, a screw head fixture (screw plate), and an M5$\times$25 Phillips head screw, as seen in~\autoref{fig:unscrewing_experimental_setup}. The cordless drill is equipped with a microcontroller that communicates with the robot controller via ROS interface.
\begin{table}[t]
\renewcommand{\arraystretch}{1.1}
\caption{Control parameters used during the different experiments.}
\begin{adjustbox}{width=\columnwidth,center}
\centering
    \begin{tabular}{l c c}
    \hline
    \hline
    \multicolumn{3}{c}{\textbf{Cartesian Impedance controller}}\\
        \hline
        Translational spring stiffness & $\Stif_{t}$ & $900\cdot\Id_3$ \\
        Rotational spring stiffness & $\Stif_{r}$ & $40\cdot\Id_3$ \\
        Coupling spring stiffness & $\Stif_{c}$ & $0\cdot\Id_3$ \\
        \hline
        \hline
        \multicolumn{3}{c}{\textbf{Hybrid Force-Impedance controller}}\\
        \hline
        Translational spring stiffness & $\Stif_{t}$ & $100\cdot\Id_3$ \\
        Rotational spring stiffness & $\Stif_{r}$ & $10\cdot\Id_3$ \\
        Coupling spring stiffness & $\Stif_{c}$ & $0\cdot\Id_3$ \\
        Desired force & $\mathbf{w}_{\text{desired},z}$ & $-\mathbf{f}_{\text{engage,z}}$\\
    \hline
    \hline
     \multicolumn{3}{c}{\textbf{Energy-Aware Impedance controller}}\\
     \hline
        Translational spring stiffness & $\Stif_{t}$ & $900\cdot\Id_3$ \\
        Rotational spring stiffness & $\Stif_{r}$ & $40\cdot\Id_3$ \\
        Coupling spring stiffness & $\Stif_{c}$ & $0\cdot\Id_3$ \\
        Max. allowed energy & $\entotUP$ & $0.7 \text{J}$\\
        Max. allowed power & $\powmotUP$ & $0.5 \text{W}$ \\
        Initial damping & $\Damp$ & $5\cdot\Id_7$\\
     \hline
     \hline
     \multicolumn{3}{c}{\textbf{Energy Tank \& Others}}\\
     \hline
     Engagement force & $\mathbf{f}_{\text{engage,z}}$ & $15 \text{N}$\\
     Max. allowed power to be extracted & $\powtankLOW$ & $-0.175 \text{W}$\\
     Max. Energy level in the tank & $\entankUP$ &$ 5 \text{J}$\\
     Min. Energy level in the tank & $\entankLOW$ & $0.5 \text{J}$\\
     Initial Energy level in the tank & $\entank$ & $3 \text{J}$\\
     \hline
     \hline
    \end{tabular}
    \end{adjustbox}
\label{tab:ctrlVar}
\vspace{-4.5mm}
\end{table}
The experiment can be divided into two different phases: \begin{enumerate*}
  \item unscrewing, and
  \item interaction with a human.
\end{enumerate*}
For our use-case, it is assumed that the position of the screw is known a priori, similarly to the work of~\cite{li2020unscrewing} who discusses an exploration method for successful tool insertion.
The unscrewing phase starts with the tool-center-point (TCP) being already inserted in the screw head and begins to increase the force along the TCPs $\Zax$ until it reaches the engagement force $\mathbf{f}_{\text{engage,z}}$. For the standard and the presented energy-aware Cartesian impedance controller, $\mathbf{f}_{\text{engage,z}}$ is achieved by moving the desired transformation $\homTr^0_d$ along the $\Zax$ in the negative direction, thereby preventing the screw bit from slipping. As soon as $\mathbf{f}_{\text{engage,z}}$ is reached, the drill starts, and simultaneously, the desired TCP pose is translated along the $\Zax$ mentioned above in a positive direction to maintain $\wrench^{0,EE}_{ext,z}$ approximately constant. With the hybrid force-impedance controller, such mechanism is ensured by the force loop of the $\Zax$.
To systematically generate contact loss, we use a simulated broken shaft, i.e., a screw $10\;\si{mm}$ shorter than expected by the robot motion planner. In this way, the controller still exerts some force while the shaft comes out from the plate. For this reason, due to $\wrench^{0,\text{EE}}_K$, the screw slips, hence losing the contact between the screw and the drill bit and the robot end-effector moves in $z$ direction towards its current $\homTr^0_d$ and eventually hits the table.    
To further evaluate the suitability of the controller in human-populated environments, after the unscrewing phase, the robot's end-effector was disturbed by a human in the interaction phase. Snapshots of the experiment are available in \autoref{fig:snapshots}.
The control parameters chosen for each control scheme are shown in Table~\ref{tab:ctrlVar}.
The force value $15\;\si{N}$ was chosen based on an initial investigation which has shown that a screw tightened with $3\;\si{Nm}$ needs approximately $15$-$20\;\si{N}$ of force applied to avoid slippage. The maximum energy threshold $\entotUP$ was selected according to ISO/TS 15066:2016~\cite{ISO} that specified a range of $0.52$ - $2.5\;\si{J}$, the value for $\powmotUP$ and the threshold for the power limit on the energy tank $\powtankLOW$ were picked according to earlier results in~\cite{hjorth2020energy,raiola2018energy} and in~\cite{ramuzat2022passive}, respectively.
\subsection{Standard Cartesian Impedance Controller}\label{sub:imp_ex}
\begin{figure}[t]
\centering
	\includegraphics[width=0.9\columnwidth,trim={0cm 0.15cm 0cm 0.},clip]{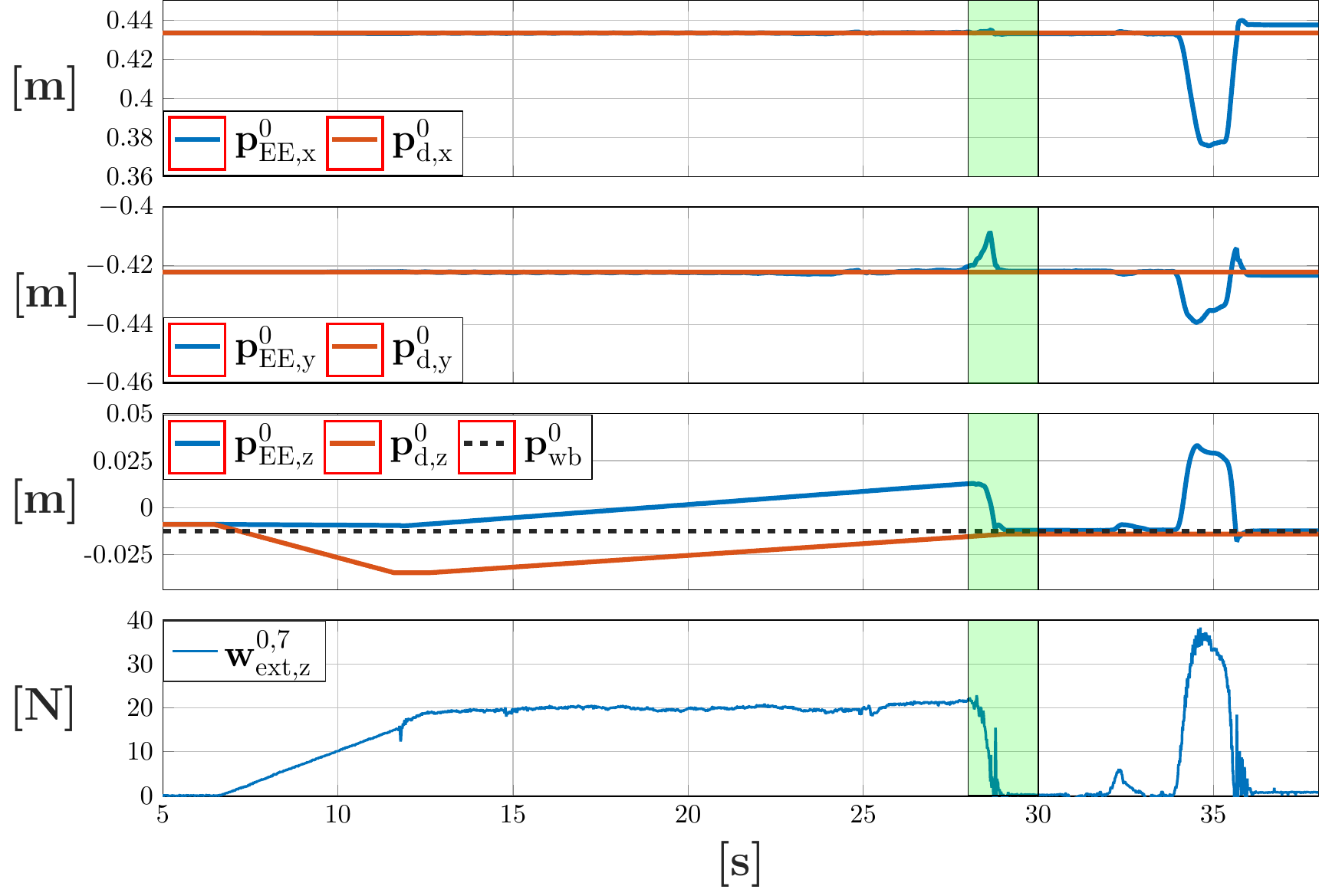}
	\caption{Visualizes the observable quantities of the Cartesian impedance controller. The difference between current $\pos^0_{\text{EE}}$ and desired position $\pos^0_{\text{d}}$ as well as the workbench position $\pos^0_{\text{wb}}$ and the linear z-component of the external force applied on the TCP $\wrench^{0,EE}_{ext,z}$.}
	\label{fig:classic_pos}
 \vspace{-4mm}
\end{figure}

\noindent As mentioned above, to generate the force required to perform the unscrewing process, $\pos^0_{d,z}$ was moved incrementally below the screw head until $\wrench^{0,\text{EE}}_{K,z}$ reaches the desired force (\autoref{tab:ctrlVar}). This change along the z-axis results in an increase in the external force applied, as seen in the time period $7-12.5\;\si{s}$ in \autoref{fig:classic_pos}.
The time period between $12.5-28\;\si{s} $ marks the unscrewing phase, where the incremental change along the z-axis is changed in a direction such that $\wrench^{0,EE}_{ext,z}$ is kept within the acceptable working range. Focusing on the external force, one can see how force decreases due to the missing counter force by the screw, which drives the robot to hit the workbench with an impact force of $15\;\si{N}$ before the equilibrium pose of the controller reaches the surface (highlighted with a green area).
In the interaction phase, which starts immediately afterwards, the robot's end effector is displaced by pHRI at $32\;\si{s}$ where the spatial spring generated a force of $37\;\si{N}$ to counteract the disturbance. Once the end effector has reached its maximal displacement of $0.34\;\si{m}$, it is released, and the robot moves back towards its desired configuration and hits the workbench a second time with $18\;\si{N}$.
\vspace{-1mm}
\subsection{Hybrid Force-Impedance Controller}\label{sub:force_ex} 
\begin{figure}[t]
\centering
	\includegraphics[width=1\columnwidth,trim={0.1cm 0.13cm 0cm 0.},clip]{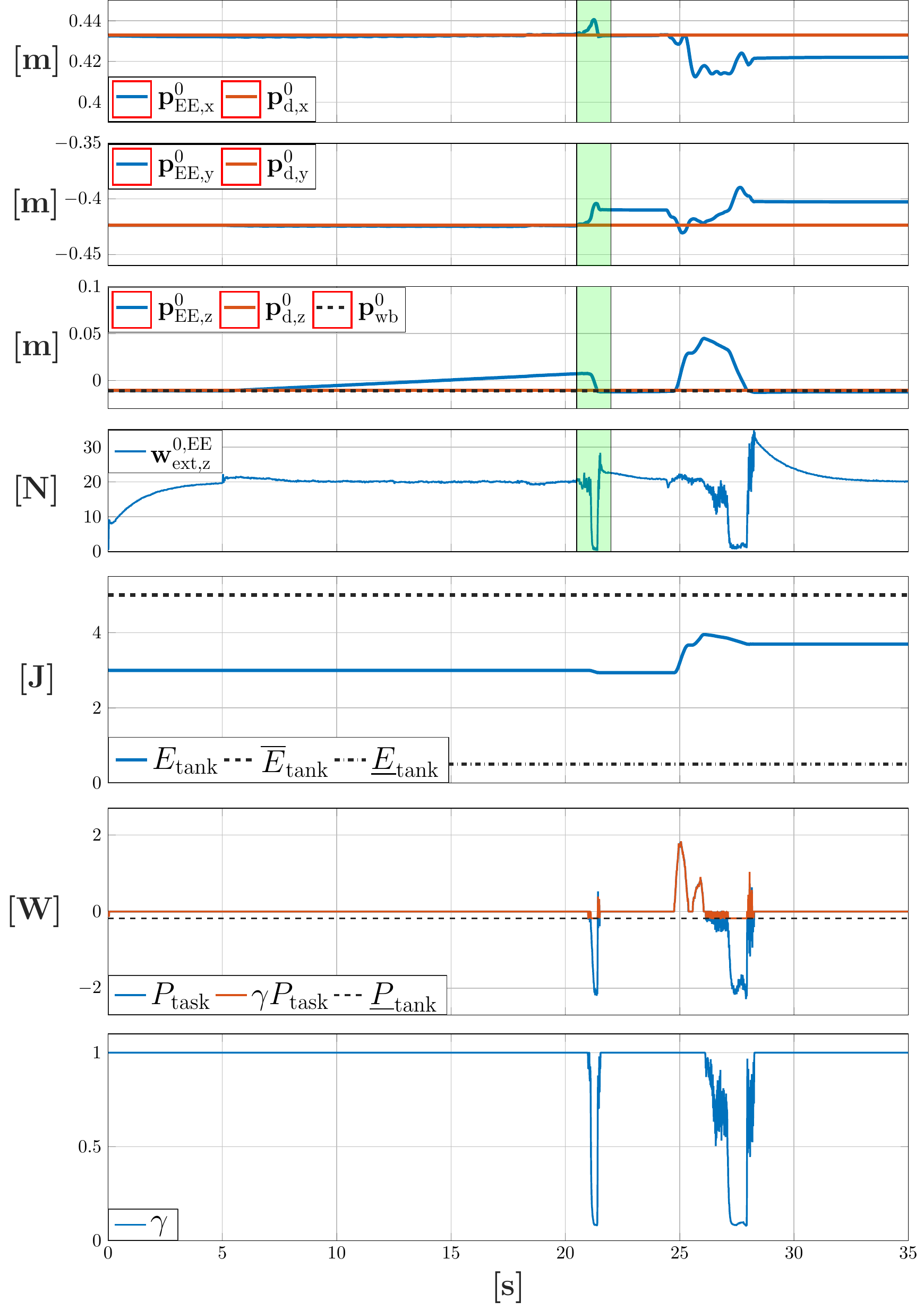}
	\caption{Visualizes the observable quantities of the hybrid force impedance controller. The difference between current $\pos^0_{\text{EE}}$ and desired position $\pos^0_{\text{d}}$ as well as the workbench position $\pos^0_{\text{wb}}$, the linear z-component of the external force applied on the TCP $\wrench^{0,EE}_{ext,z}$ and the energy tank $\entank$ and its power flow $\powtask$. }
	\label{fig:force_force}
\vspace{-4mm}
\end{figure}

\noindent In the case of the hybrid controller, the force that is necessary to enable the the screw to be unscrewed is generated by the force control part. Therefore, it is not necessary to manipulate the desired pose $\homTr^0_d$, as seen in \autoref{fig:force_force}. 
After $5\;\si{s}$, the desired force of $20\;\si{N}$ is reached and the unscrewing phase begins, which lasts until $21\;\si{s}$.
After the contact loss, the robot gets accelerated by the force controller in order to maintain the desired force. Even though the restrictions on the implemented energy tank already reduce the rate by which the controller can inject energy into the system (see ~\autoref{fig:force_force}), the resulting force is still significant (approx. $26\;\si{N}$). In the interaction phase, which begins immediately afterwards, the robot end-effector is displaced by pHRI at $24.5\;\si{s}$. During the displacement of the robot, one can see how the force response tries to adapt to the applied force. When the end effector is released, the tank's power limit takes effect, thereby reducing the force response; however, even though the robot makes contact with the work surface with a significant impact force (approx. $34\;\si{N}$).
\subsection{Energy-Aware Controller}\label{sub:energ_aware_ex}
\begin{figure}[t]
\centering
	\includegraphics[width=0.9\columnwidth,trim={0cm 0.125cm 0cm 0.},clip]{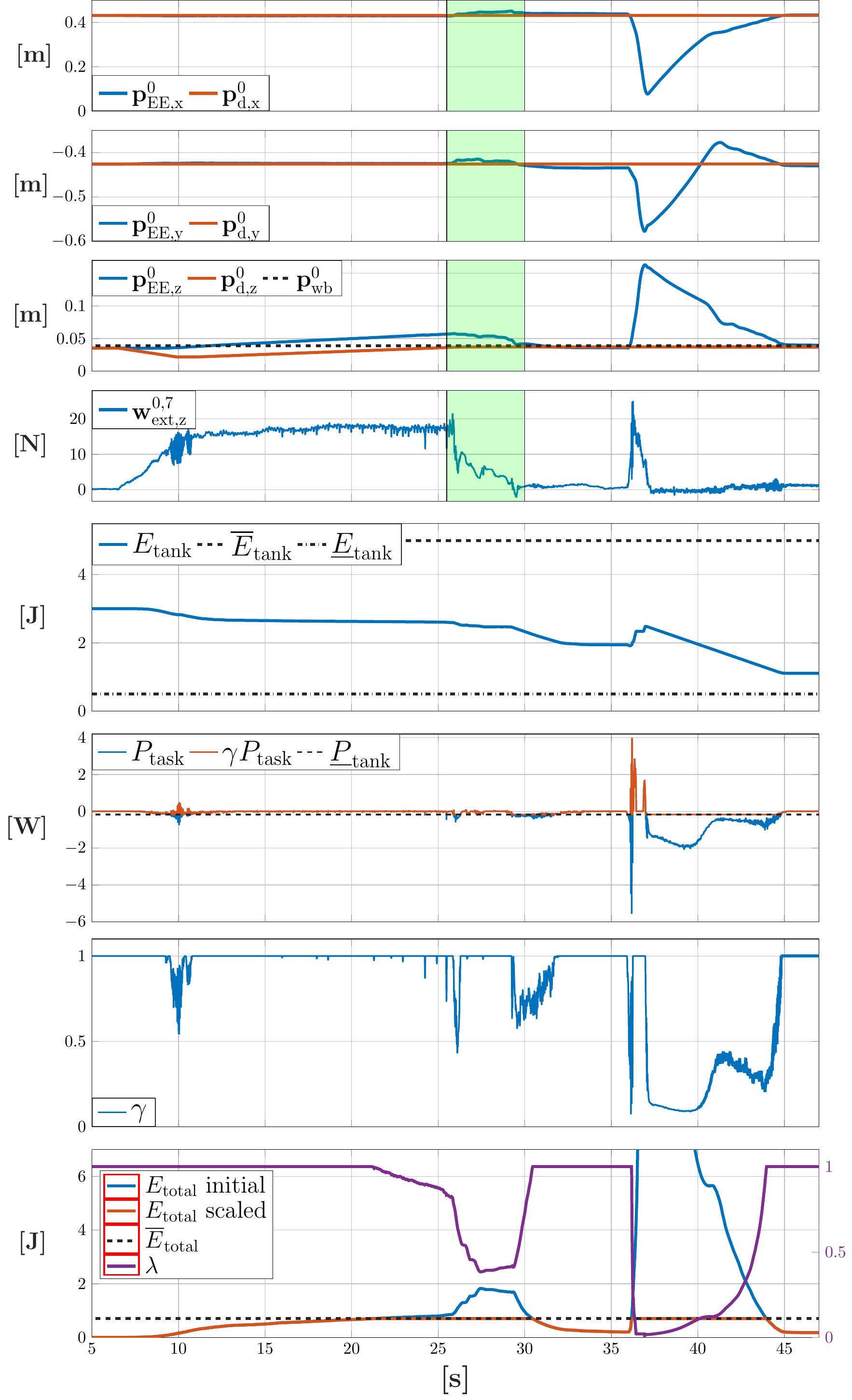}
	\caption{Visualizes the observable quantities of the energy-aware impedance controller. The difference between current $\pos^0_{\text{EE}}$ and desired position $\pos^0_{\text{d}}$, as well as the workbench position $\pos^0_{\text{wb}}$, the linear z-component of the external force applied on the TCP $\wrench^{0,EE}_{ext,z}$, the energy tank $\entank$ and its power flow $\powtask$ as well as the energy scaling of the total energy $\entot$.}
	\label{fig:energy_force}
 \vspace{-6mm}
\end{figure}

\noindent The unscrewing task here follows the same procedure as the standard Cartesian impedance controller, where the equilibrium's pose has to be incrementally moved along the screw's longitudinal axis. The desired force is reached after $10\;\si{s}$ (\autoref{fig:energy_force}), and the equilibrium pose is again translated in an upward direction during the unscrewing phase.
At the point of contact loss ($26\;\si{s}$), the end effector starts to move towards its equilibrium pose thereby making contact with the workbench at $29\;\si{s}$ without a significant impact force (approx. $1.2\;\si{N}$).
A similar behavior can be observed during the interaction phase, where the end effector hits the workbench after the end of the disturbance without resulting in a significant impact force (approx. $2.6\;\si{N}$).
This can be attributed to the effect of the spring energy scaling and the power flow regulation based on the power limit in the energy tank. When $\lambda$ decreases, the stiffness of the spring also decreases and thereby reduces the energy in the system; however, when $\lambda$ increases again, energy is injected back into the system; if this injection is not kept in check, it can result in unstable and non-passive behavior. Therefore, the rate at which energy can be injected must be monitored and limited. When comparing the response of $\lambda$ and $\gamma$, one can see that at every instant when $\lambda$ increases, $\gamma$ decreases. As in \eqref{eq:tau_motion_new}, $\gamma$ directly affects the rate at which energy can be injected into the system, that is, it regulates how fast $\lambda$ can increase at each time step. Additionally, the energy tank also ensures that $\lambda$ can only be increased if there is energy in the tank ($\entank>\entankLOW$). If the energy tank is depleted, $\lambda$ remains constant.

\section{Discussion~\&~Conclusions}
\label{sec:Conclusion}
\noindent This work presents an energy-based control formalism in combination with an augmented energy tank to ensure the passivity of the system. To the best of the authors knowledge, such a formalism has not been applied to a disassembly task before. The capabilities of the formalism to handle contact loss and pHRI are evaluated on an unscrewing task. Additionally, its performance was compared to a standard Cartesian impedance controller as well as a hybrid force-impedance controller. When it comes to applying and tracking a constant force onto the screw, the force-impedance control outperforms the standard Cartesian impedance controller and the energy-aware controller; however, in case of a contact loss, it generates the highest impact force among the compared controllers. On the other hand, the Cartesian impedance controller is intrinsically passive, but, due to the task requirements in terms of force and, thus, stiff behavior, a significant impact force is generated, which is not in compliance with pHRI. The presented formalism introduces an energy-aware scaling mechanism to the Cartesian impedance controller, as well as a power flow regulated energy tank, which ensures the passivity of the system. Therefore, to allow the manipulator to react in a compliant manner in the scenario of contact loss or pHRI, the manipulator stays within its predefined energy thresholds.
Notably, during the pHRI, a larger displacement could be achieved whilst reducing the impact force between the tool and the robot (up to $92\%$), given the same initial end-effector impedance values as the classic Cartesian impedance controller.
Moreover, when facing comparable disturbances, such as in the contact loss scenario, the impact force is reduced by up to $91\%$ and $95\%$ for the Cartesian Impedance and the hybrid force impedance controller respectively. 
One drawback of the presented control formalism is the parameterization of the power flow limit, i.e. the size of the energy tank must be designed manually. In the direction, in~\cite{michel2022safety} the power flow formulation depends on the remaining energy in the tank.

\balance
\bibliographystyle{IEEEtran}
\bibliography{ICRA_bib.bib}
\end{document}